\documentclass{article}

\PassOptionsToPackage{numbers,comma,compress}{natbib}

\usepackage[final]{neurips_mlsb_2022}

\usepackage[algo2e,ruled,noend]{algorithm2e}
\usepackage[utf8]{inputenc} 
\usepackage[T1]{fontenc}    
\usepackage{url}            
\usepackage{booktabs}       
\usepackage{microtype}      
\usepackage{xcolor}         
\usepackage{enumitem}
\usepackage{tikz}
\usepackage{pgfplots}
\pgfplotsset{compat=1.17} 
\usepackage{pgfplotstable}
\usepackage{multirow}
\usepackage[para,online,flushleft]{threeparttable}
\usepackage[labelfont=bf,textfont=it]{caption}
\usepackage{subcaption}
\usepackage{titlesec}
\usetikzlibrary{spy}

\usepackage{textgreek}
\usepackage{comment}

\usepackage{tabulary}

\titlespacing\section{0pt}{3pt plus 1pt minus 1pt}{3pt plus 1pt minus 1pt}
\titlespacing\subsection{0pt}{2pt plus 1pt minus 1pt}{2pt plus 1pt minus 1pt}
\titlespacing\subsubsection{0pt}{2pt plus 2pt minus 2pt}{2pt plus 2pt minus 2pt}
\titlespacing\paragraph{0pt}{1pt plus 2pt minus 1pt}{1pt plus 2pt minus 1pt}

\usepackage{amsmath,amsfonts,bm, amsthm}
\usepackage{algorithm}
\usepackage[noend]{algorithmic}
\usepackage{nicefrac}        
\usepackage{chngcntr}

\newcommand{\colref}[2]{\hyperref[#2]{#1~\ref*{#2}}}
\newcommand{\coloredref}[2]{\hyperref[#2]{#1~\ref*{#2}}}
\newcommand{\coloredsubref}[3]{\hyperref[#2]{#1~\ref*{#2}{#3}}}

\newcommand{\Figref}[1]{\colref{Figure}{#1}}

\newcommand{\Secref}[1]{\colref{Section}{#1}}

\def\eqref#1{\colref{Equation}{#1}}

\newcommand{\Tabref}[1]{\colref{Table}{#1}}

\def\1{\bm{1}}

\DeclareMathAlphabet{\mathsfit}{\encodingdefault}{\sfdefault}{m}{sl}
\SetMathAlphabet{\mathsfit}{bold}{\encodingdefault}{\sfdefault}{bx}{n}

\theoremstyle{plain}

\theoremstyle{remark}

\theoremstyle{definition}

\theoremstyle{plain}

\theoremstyle{plain}

\theoremstyle{definition}

\providecommand{\corollaryname}{Corollary}
\providecommand{\lemmaname}{Lemma}
\providecommand{\problemname}{Problem}
\providecommand{\remarkname}{Remark}
\providecommand{\theoremname}{Theorem}

\counterwithin*{theorem}{section}

\usepackage{hyperref}       
\hypersetup{
    colorlinks=true,
    linkcolor={blue},
    citecolor={blue},
    urlcolor={blue},
    breaklinks=true,
	plainpages=true
}
\theoremstyle{plain}

\theoremstyle{definition}

\theoremstyle{remark}

\title{3D Reconstruction of Protein Complex Structures Using Synthesized Multi-View AFM Images}

\author{
    Jaydeep Rade\textsuperscript{1} \\
    \And
    Soumik Sarkar\textsuperscript{1} \\
    \And
    Anwesha Sarkar\textsuperscript{1} \\
    \And
    Adarsh Krishnamurthy\textsuperscript{1} \\
    \And
    Iowa State University\textsuperscript{1}\\
    \texttt{\{jrrade, soumiks, anweshas, adarsh\}@iastate.edu}\\
    \\
}

\begin{document}

\maketitle

\begin{abstract}
Recent developments in deep learning-based methods demonstrated its potential to predict the 3D protein structures using inputs such as protein sequences, Cryo-Electron microscopy (Cryo-EM) images of proteins, etc. However, these methods struggle to predict the protein complexes (PC), structures with more than one protein. In this work, we explore the atomic force microscope (AFM) assisted deep learning-based methods to predict the 3D structure of PCs. The images produced by AFM capture the protein structure in different and random orientations. These multi-view images can help train the neural network to predict the 3D structure of protein complexes. However, obtaining the dataset of actual AFM images is time-consuming and not a pragmatic task. We propose a virtual AFM imaging pipeline that takes a 'PDB' protein file and generates multi-view 2D virtual AFM images using volume rendering techniques. With this, we created a dataset of around 8K proteins. We train a neural network for 3D reconstruction called Pix2Vox++ using the synthesized multi-view 2D AFM images dataset. We compare the predicted structure obtained using a different number of views and get the intersection over union (IoU) value of 0.92 on the training dataset and 0.52 on the validation dataset. We believe this approach will lead to better prediction of the structure of protein complexes.

\end{abstract}

\section{Introduction}

Accurate prediction of a 3D protein structure based on its amino acid sequence has been a challenging research task for more than a five decades~\citep{PF2008Dill} and is fundamentally important since the structure defines the functionality of proteins~\citep{PF20212Dill}. Researchers have been utilizing and improving experimental methods like x-ray crystallography, nuclear magnetic resonance (NMR) spectroscopy, and cryo-electron microscopy (cryo-EM) to determine the 3D structures of proteins~\citep{NMR2001Kurt, Thompson2020AdvancesIM,cryoEM2015Bai, Jasklski2014ABH}. Improvements in experimental methods have increased the availability of structural data for a large number of proteins. In parallel, the advancements in deep learning (DL) have had a significant impact on protein structure prediction, with the development of AlphaFold2~\citep{AF22021Jumper} and RoseTTAFold~\citep{RF2021Baek}. Combining DL with advancements in cryo-EM imaging of single particle analysis enabled the determination of protein structure and complexes. However, Cryo-EM images can be very noisy, with a signal-to-noise ratio as low as $-20 dB$~\citep{Gupta2020MultiCryoGANRO}. In addition, the Cryo-EM experiments are expensive and can cost up to \$1500 per structure~\citep{Cianfroco2015}. 

\begin{figure}[t!]
    \centering
    \includegraphics[width=0.8\linewidth, trim={4in 3.3in 4in 2.75in},clip]{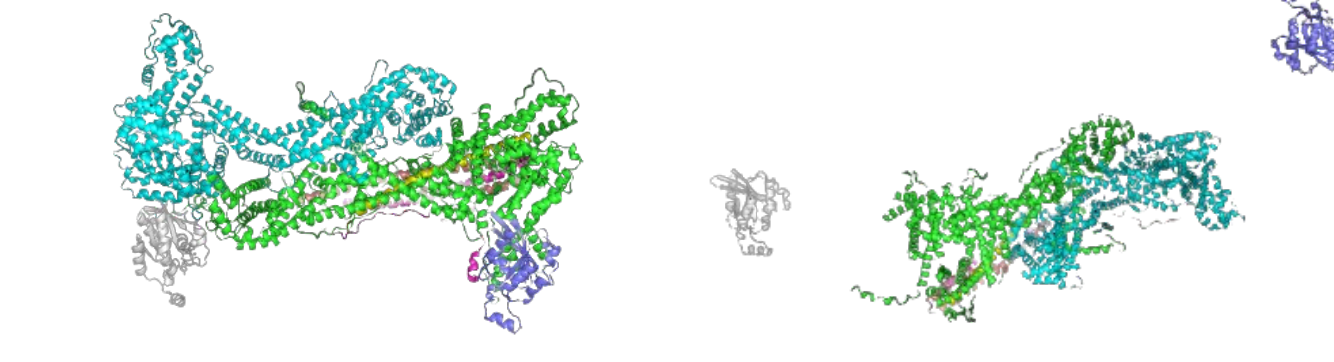}
    \caption{AlphaFold2 multimer (right) fails to predict the structure of WRC protein complex (left).}
    \label{fig:AlphaFold_prediction}
\end{figure}

Another way to infer the structure of proteins is to learn the mapping from a protein's sequence of amino acids to its structure. These methods include the use of Position Specific Scoring Matrices (PSSM)~\citep{Guo2014}, which compute the single residue mutation frequency distribution, and Multiple Sequence Alignment (MSA), which determines the co-evolutionary information~\citep{Senior2020ImprovedPS, Yang2020}. AlphaFold2 uses entire MSAs and a template-directed inference process. However, AlphaFold2 is not as effective in predicting the structure of protein complexes where proteins are bonded together. For example AlphaFold2-multimer~\citep{AF2Multi2022Evans}, which is trained specifically for protein complex structure prediction, fails to predict the structure of the WRC protein complex~\citep{koronakis2011wave,rottner2021wave,ismail2009wave,chen2014wave} as seen in \Figref{fig:AlphaFold_prediction}. As \textit{In Silico} methods are unable to infer the structure of protein complexes in certain cases, and Cryo-EM methods can be expensive and time-consuming, we propose to train the DL model with images captured using Atomic Force Microscopy (AFM). AFM is a part of scanning probe microscopy, a non-invasive technique that provides the platform for high-resolution imaging of proteins in their physiological environment without prolonged, complicated sample preparation techniques (freezing, drying, dye-tagging) that ultimately damage these soft samples.

With these advantages of AFM, we propose a novel approach to infer the 3D structure of protein complexes using a neural network that is trained using multi-view AFM images. We use Pix2Vox++~\citep{pix2vox2020Xie} neural network architecture for predicting the 3D protein structure. We explain the details of the architecture in \Secref{sec:methods}. Even though AFM is not as expensive as Cryo-EM, experimentally acquiring the AFM image dataset for training is not a practical solution. Hence we develop a virtual AFM imaging pipeline to generate multiple views of the protein complexes using a volume rendering technique. We discuss this virtual imaging process in detail in \Secref{sec:dataset}. We show our preliminary results in \Secref{sec:results}. Finally, we conclude our work with some future direction in \Secref{sec:conclusion}.

\section{Method}\label{sec:methods}

\begin{figure}[b!]
    \centering
    \includegraphics[width=0.99\linewidth, trim={0.5in 1.5in 0.5in 1.5in},clip]{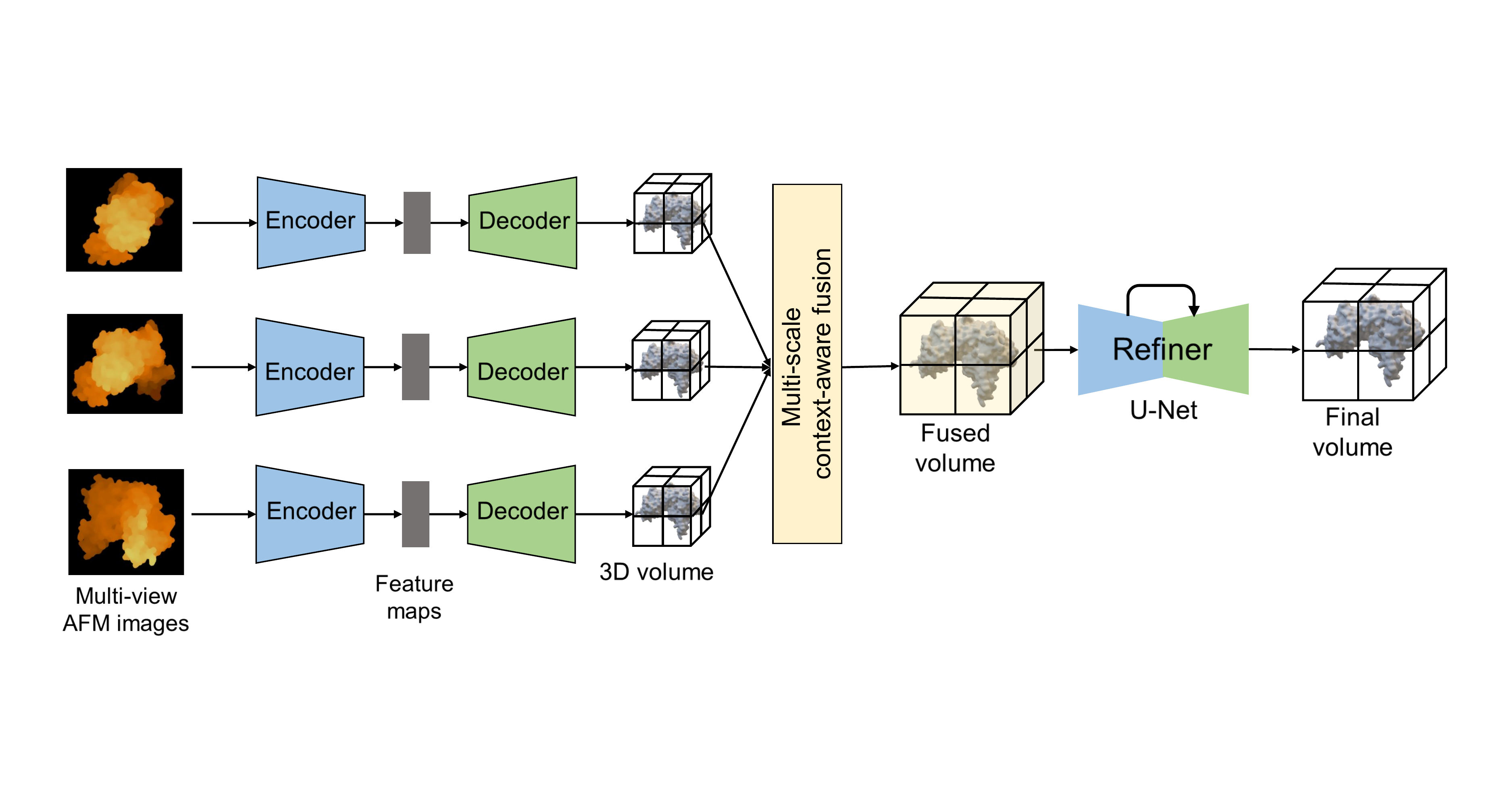}
    \caption{The architecture of Pix2Vox++ neural network which reconstructs the 3D shape from multi-view 2D images.}
    \label{fig:architecture}
\end{figure}

We implement the Pix2Vox++ neural network architecture, which was proposed for the 3D reconstruction of the shape of an object using single or multiple images. The goal is to predict the 3D shape represented in the voxel occupancy grid where 0 value indicates the empty voxel and 1 indicates the occupied voxel. It includes four blocks: encoder, decoder, multi-scale context-aware merger, and a refiner. The overall forward pass is summarized in \Figref{fig:architecture}. The encoder takes each input image and transforms it into latent space feature maps. Next, the decoder processes each latent feature map and generates a 3D volume for each input image. These generated 3D volumes are fused into one using a multi-scale context-aware fusion block. It adaptively assigns the weight for each voxel in all 3D volumes to select high-quality reconstruction. In the end, the refiner further improves the reconstruction by rectifying the inaccurate parts in generated fused 3D shape and outputs the final 3D shape.

The encoder uses 2D convolutional layers to transform input images into latent representation feature maps, which decoders will use to generate 3D volume per image. Using 2D convolutional layer followed by non-linearity ReLU function, batch-normalization, and then max-pooling operation to encode the image into $4\times4\times256$ feature map. The feature map is then reshaped to $2\times2\times2\times512$ and fed into the 3D transposed convolutional layer, followed by the ReLU function, which up-samples the feature map into a 3D volume for each input image. As the inputs are multi-view images, reconstruction from the visible parts is better than the invisible parts, and this observation can be used to adaptively select the high-quality parts from each generated volume. Multi-scale context-aware fusion block combines these selected high-quality parts into a single 3D volume. This block uses convolutional layers with non-linearity functions to generate the score per voxel per 3D volume, a higher score for better quality reconstruction. Finally, the refiner tries to correct the incorrect reconstruction in the fused volume. The refiner uses a network similar to U-Net~\citep{ronneberger2015u} to output the final 3D shape.    

\section{Dataset}\label{sec:dataset}

\begin{figure}[t!]
    \centering
    \includegraphics[width=0.8\linewidth, trim={0.5in 0.5in 0.5in 0.5in},clip]{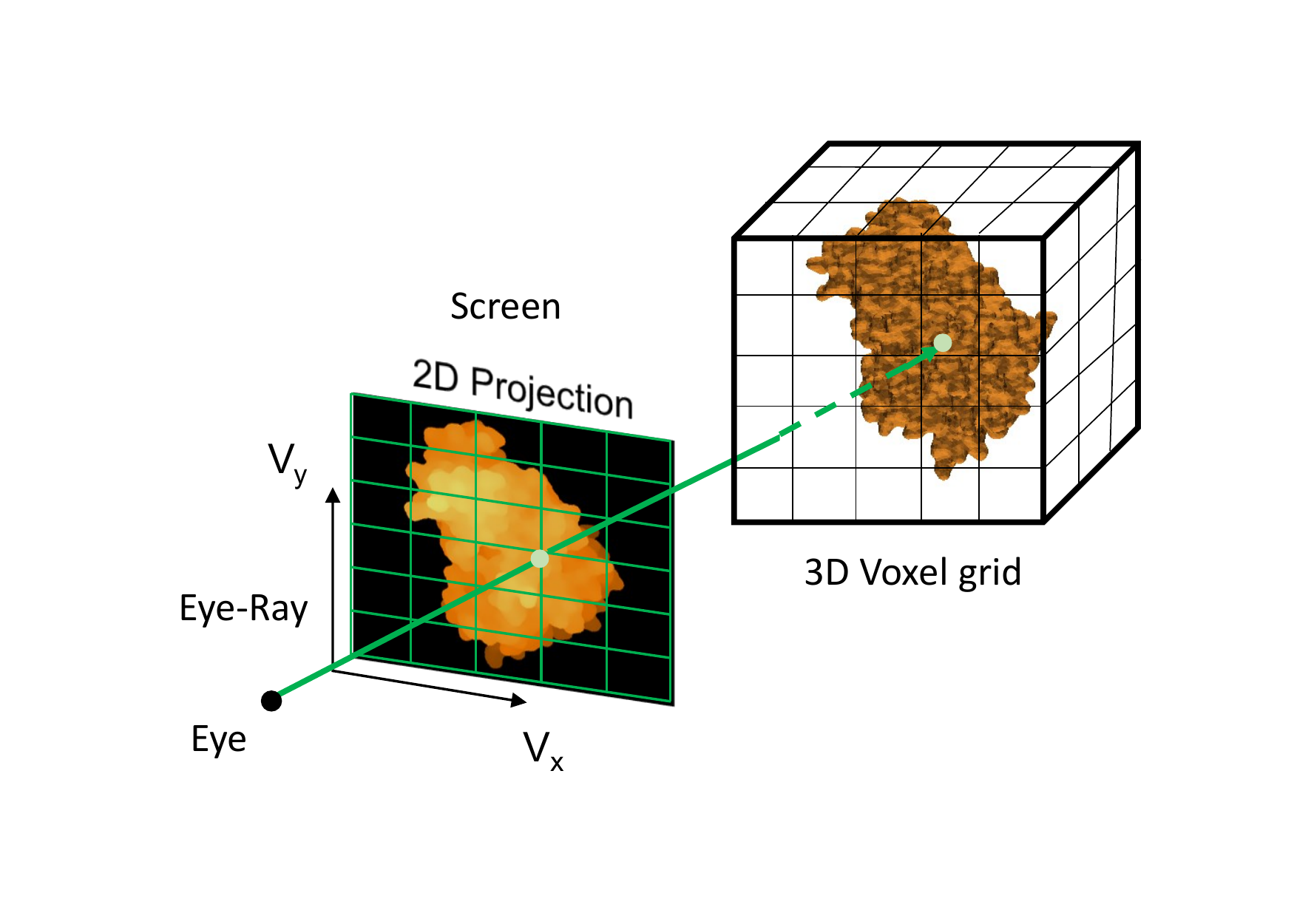}
    \caption{Volume rendering of 3D voxelized protein structure to obtain AFM-like 2D images.}
    \label{fig:volumeRendering}
\end{figure}

We have developed the pipeline to generate the virtual multi-view AFM images from a 'PDB' format protein file. First, we download the 'PDB' files of proteins from the Protein Data Bank. Using PyMol software often used to visualize proteins, we transform the 'PDB' format structure into a 3D mesh 'OBJ' file format. Proteins can be visualized in various representations such as surfaces, spheres, sticks, cartoons, or ribbons. We export the 'OBJ' files for surface representations. We then use GPU-accelerated algorithm~\citep{YOUNG201811} to voxelized the 3D mesh. The resolution of the voxelization is set to 256 voxels along the longest dimension, and then zero padding is performed along the remaining dimensions, so the final voxelized structure has the resolution of $256\times256\times256$. For the current work, we down-sample this high-resolution structure to a $32\times32\times32$ grid. 

The GPU-accelerated volume rendering technique is used to get the 2D projections of the voxelized 3D protein structure as shown in \Figref{fig:volumeRendering} to replicate the AFM imaging task. We rotate the 3D structure randomly in 3D space to obtain the multi-view images. A total of 25 multi-view images are saved using an image format. Using this data pipeline, we construct the dataset of around 8K protein samples containing the pairs of voxelized structures and corresponding 25 multi-view virtual AFM images. Additionally, we performed high-resolution (nanometer scale) imaging of the single molecules of the WRC protein-complex~\citep{koronakis2011wave,rottner2021wave,ismail2009wave,chen2014wave,Chen2022WRC}. We also generate the virtual AFM images of the WRC complex. These WRC samples, both actual and synthesized images, will be the testing samples instead of testing the trained network on the actual AFM images.

\section{Training Details}
We split the dataset into training and validation datasets. The $80\%$ of the data forms the training set, and the remaining $20\%$ forms the validation set. As we have generated $25$ random views for each protein structure, we can experiment with a different number of views as input to the network. To increase the number of samples, we repeat the samples four times with different views as inputs but the same 3D structure as output. The input RGB image resolution used is $224\times224\times3$ and the output resolution is $32\times32\times32$. The total number of parameters in the neural network is around $58.9Mn$. A binary cross-entropy (BCE) loss function is used as we are predicting the occupancy 3D voxel grid. We use a batch size of $32$ and train for $150$ epochs. We use the learning rate of $0.001$ with the Adam optimizer~\citep{kingma2014adam}. We used NVIDIA A-100 GPUs for running all the experiments.

\section{Results and Discussion}\label{sec:results}

We compute the intersection over union (IoU) metric between ground truth and predicted 3D structure; the value is between 0 and 1, while 0 denotes the no overlap and 1 denotes the perfect overlap. We evaluate the IoU value on training, validation data set, and test data with actual AFM images. We also compare the results using a different number of views. We summarize these experiments and corresponding metrics in \Tabref{tab:results}. We observe a significant difference between training and validation metrics, which shows that we are overfitting. We tried other techniques to overcome the overfitting issue, such as using L-2 regularization, different learning rates, and a learning rate scheduler, which didn't help to avoid the problem.

\begin{table}[h!]
  \caption{Comparing the IoU metrics for different number of views.}
  \label{tab:results}
  \centering
  \begin{tabular}{cccc}
    \cmidrule(r){1-4}
    $\#$views & Train IoU & Validation IoU & Test IoU \\
    \midrule
    3 & 0.89 & 0.50 & 0.19 \\
    5 & 0.91 & 0.51 & 0.24 \\
    8  & 0.92  & 0.51 & 0.20  \\
    10 & 0.92 & 0.52 & 0.22  \\
    \bottomrule
  \end{tabular}
\end{table}


\begin{figure}[b!]
    \centering
    \includegraphics[width=0.9\linewidth, trim={2.5in 1.5in 2.5in 1.9in},clip]{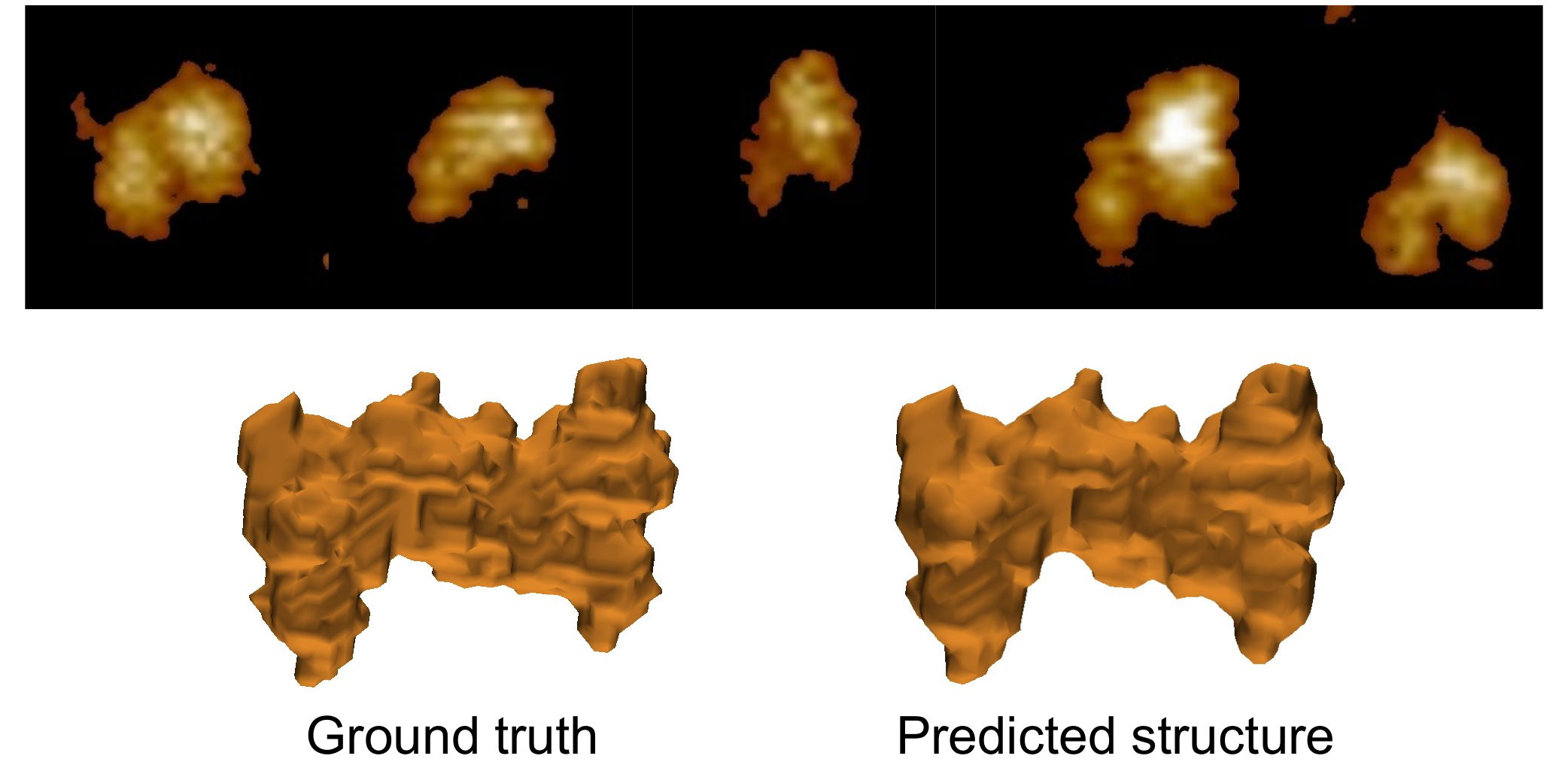}
    \caption{Comparing the visualization of ground truth and predicted structure using actual AFM images. We train the network with the actual AFM images added to the training dataset.}
    \label{fig:WRAC_pred}
\end{figure}

\begin{figure}[t!]
    \centering
    \includegraphics[width=0.99\linewidth, trim={0.25in 1.25in 0.25in 1.25in},clip]{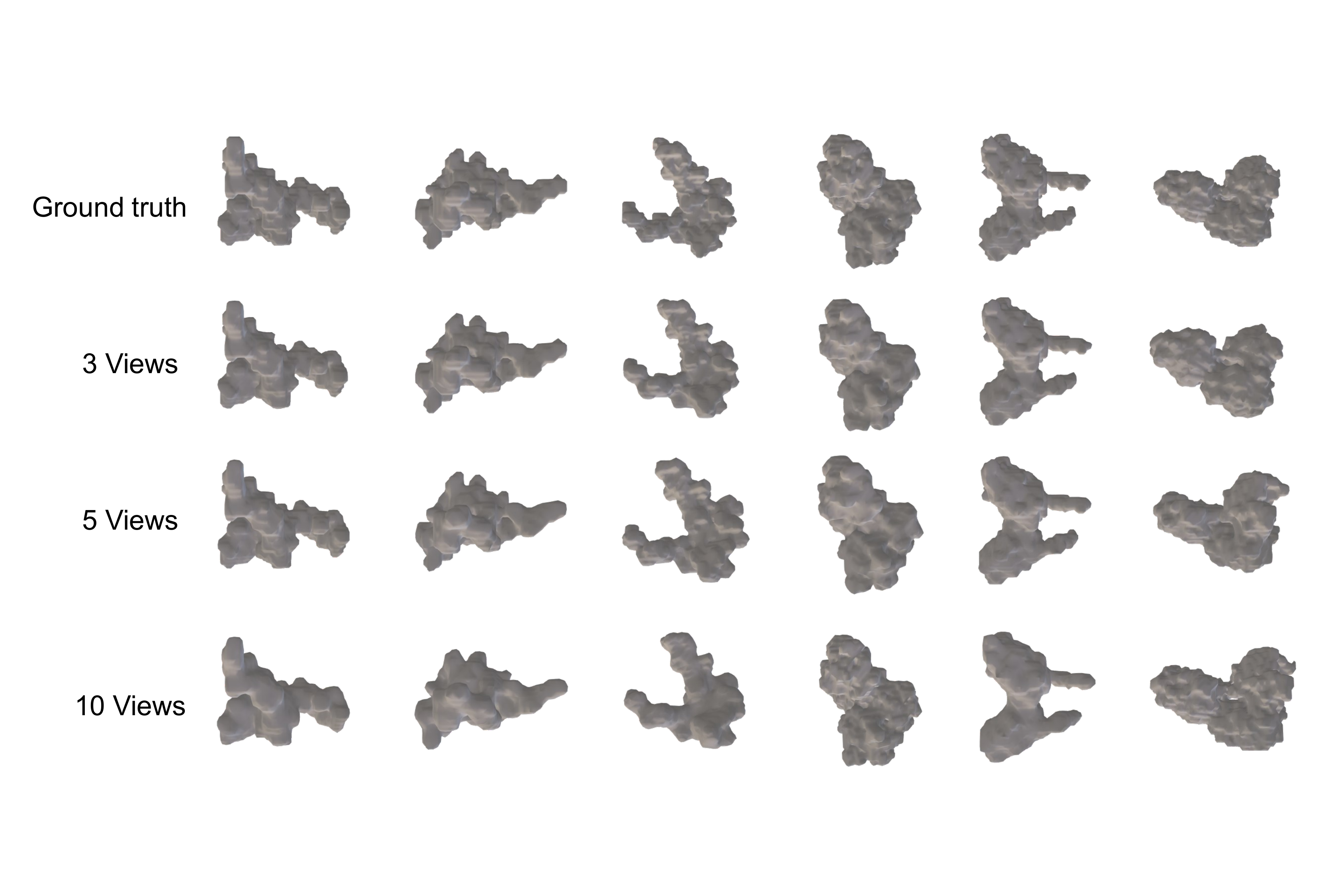}
    \caption{Predictions on samples from training dataset.}
    \label{fig:pred_train}
\end{figure}

\begin{figure}[ht!]
    \centering
    \includegraphics[width=0.99\linewidth, trim={0.25in 1.5in 0.25in 1.5in},clip]{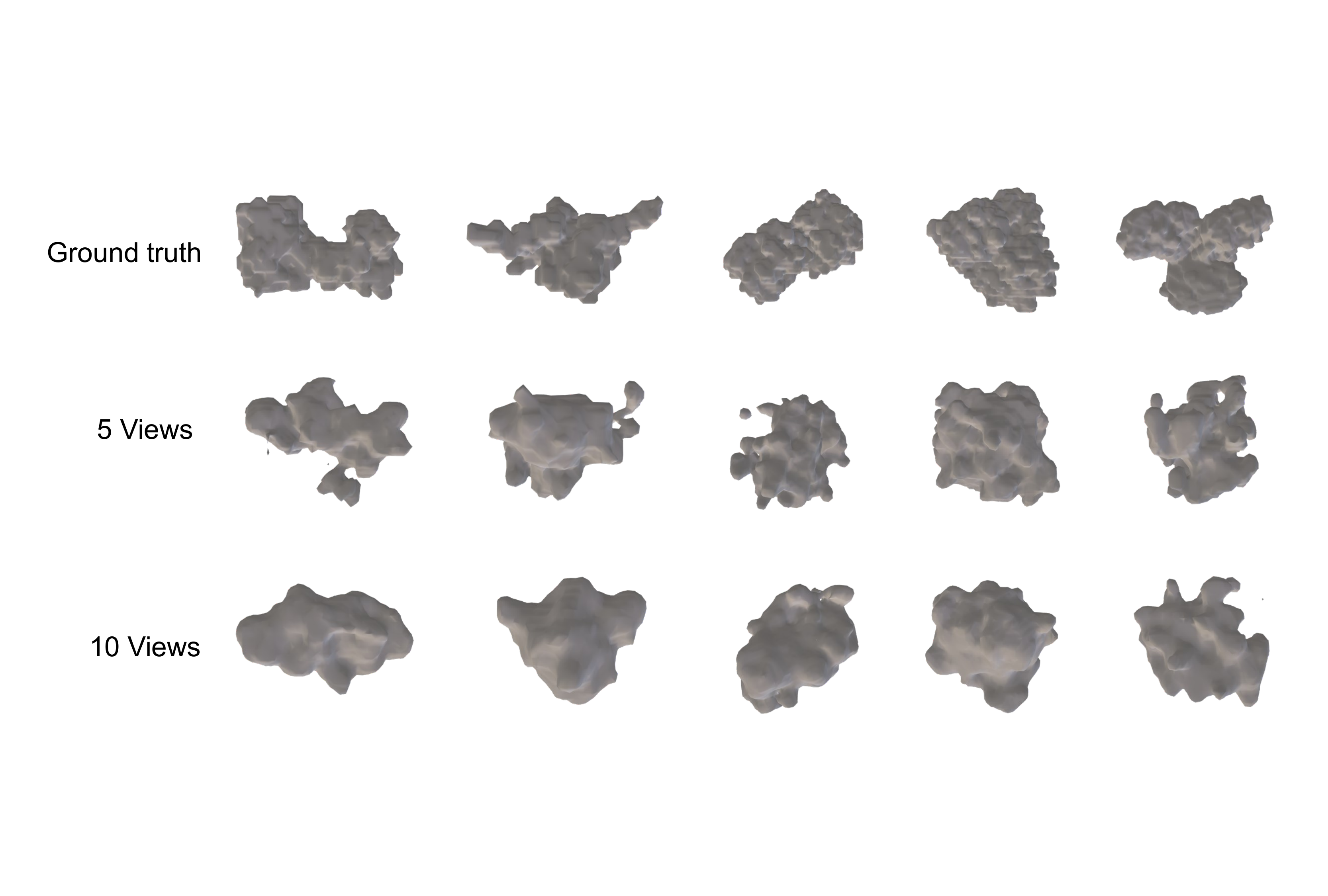}
    \caption{Predictions on samples from validation dataset.}
    \label{fig:pred_val}
\end{figure}

We tried training the network with the actual images added to the training set to determine if the network could learn from actual AFM images. We visualize the prediction and compare it to the ground truth structure in \Figref{fig:WRAC_pred}. We also show some anecdotal samples from the training and validation set in \Figref{fig:pred_train} and \Figref{fig:pred_val}, respectively.

\section{Conclusions}\label{sec:conclusion}
In this work, we propose to use AFM imaging-based deep learning 3D reconstruction approach to predict the 3D structure of protein complexes. We utilized the GPU-accelerated volume rendering technique to replicate the AFM imaging process through which we can generate the virtual AFM-like images from the ``PDB'' formal file of proteins. We generated 8K protein samples and trained the Pix2Vox++ architecture for 3D reconstruction using multi-view AFM images. Although we are facing the problem of overfitting, we could validate the approach by adding actual AFM images into the training data set and predicting the structure of the WRC protein complex. In future work, we will generate more data samples. We plan to provide additional input to the neural network, such as protein sequence embedding using large language models trained for protein sequences.

\vfill
\pagebreak
{
\small
\bibliographystyle{unsrtnat}
\bibliography{references}
}

\vfill
\pagebreak


\section*{Checklist}


\begin{enumerate}

\item For all authors...
\begin{enumerate}
  \item Do the main claims made in the abstract and introduction accurately reflect the paper's contributions and scope?
    \answerYes{}
  \item Did you describe the limitations of your work?
    \answerYes{}
  \item Did you discuss any potential negative societal impacts of your work?
    \answerNA{}{}
  \item Have you read the ethics review guidelines and ensured that your paper conforms to them?
    \answerYes{}
\end{enumerate}

\item If you are including theoretical results...
\begin{enumerate}
  \item Did you state the full set of assumptions of all theoretical results?
    \answerNA{}
        \item Did you include complete proofs of all theoretical results?
    \answerNA{}
\end{enumerate}

\item If you ran experiments...
\begin{enumerate}
  \item Did you include the code, data, and instructions needed to reproduce the main experimental results (either in the supplemental material or as a URL)?
    \answerNo{}
  \item Did you specify all the training details (e.g., data splits, hyperparameters, how they were chosen)?
    \answerYes{}
        \item Did you report error bars (e.g., with respect to the random seed after running experiments multiple times)?
    \answerNA{}
        \item Did you include the total amount of compute and the type of resources used (e.g., type of GPUs, internal cluster, or cloud provider)?
    \answerYes{}
\end{enumerate}

\item If you are using existing assets (e.g., code, data, models) or curating/releasing new assets...
\begin{enumerate}
  \item If your work uses existing assets, did you cite the creators?
    \answerNA{}
  \item Did you mention the license of the assets?
    \answerNA{}
  \item Did you include any new assets either in the supplemental material or as a URL?
    \answerNA{}
  \item Did you discuss whether and how consent was obtained from people whose data you're using/curating?
    \answerNA{}
  \item Did you discuss whether the data you are using/curating contains personally identifiable information or offensive content?
    \answerNA{}
\end{enumerate}

\item If you used crowdsourcing or conducted research with human subjects...
\begin{enumerate}
  \item Did you include the full text of instructions given to participants and screenshots, if applicable?
    \answerNA{}
  \item Did you describe any potential participant risks, with links to Institutional Review Board (IRB) approvals, if applicable?
    \answerNA{}
  \item Did you include the estimated hourly wage paid to participants and the total amount spent on participant compensation?
    \answerNA{}
\end{enumerate}

\end{enumerate}

\vfill
\pagebreak

\appendix
\setcounter{page}{1}
\section*{Appendix}

\section{Examples of Virtual AFM images}
We show some examples of generated virtual multi-view AFM images. We also compare the actual and virtual AFM images of WRC protein complex. 

\begin{figure}[ht!]
    \centering
    \includegraphics[width=0.95\linewidth, trim={2.5in 1.5in 2.5in 1.5in},clip]{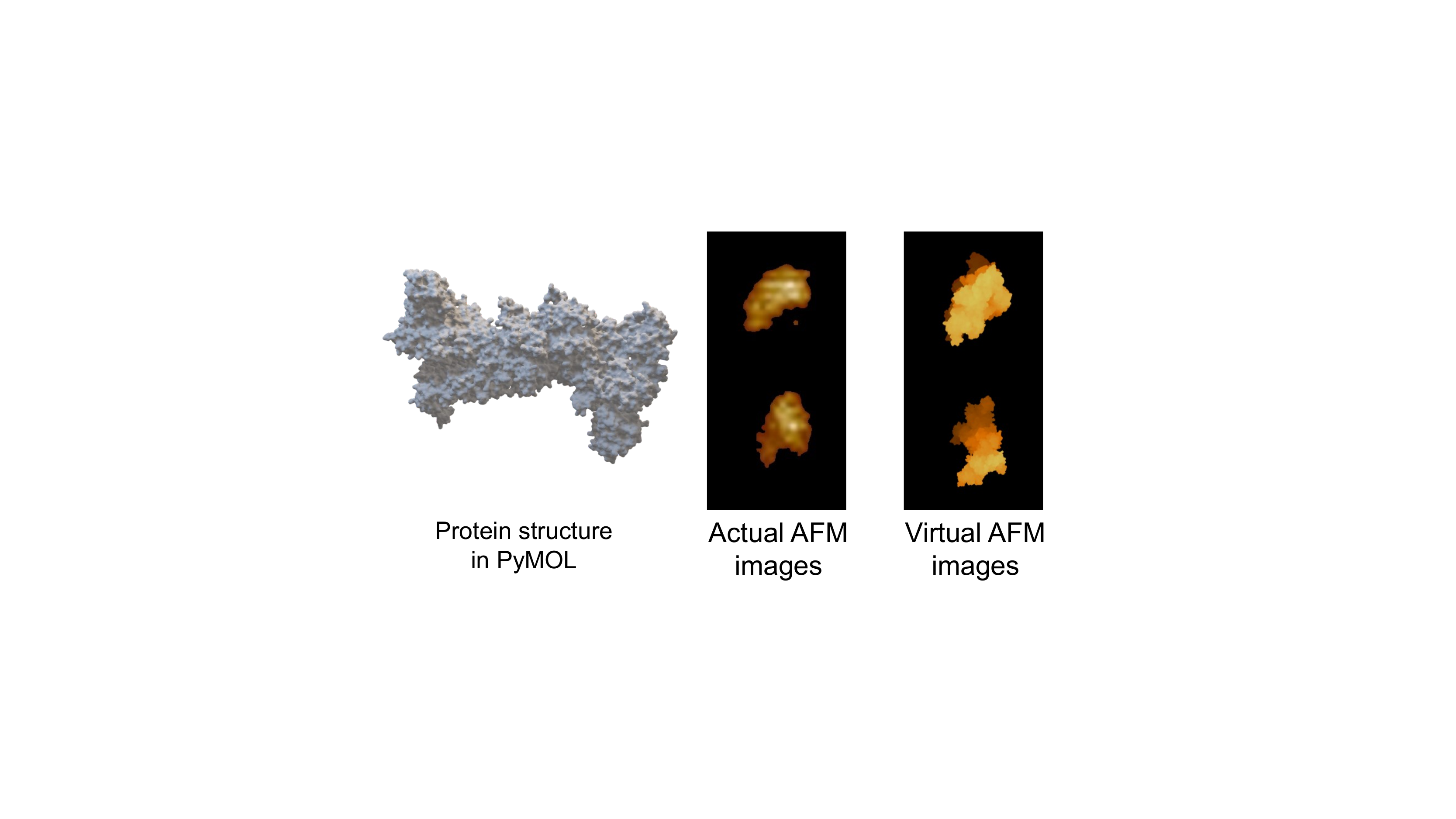}
    \caption{Comparing the virtual and actual AFM images.}
    \label{fig:virtual_vs_actual}
\end{figure}

\begin{figure}[ht!]
    \centering
    \includegraphics[width=0.95\linewidth, trim={2.0in 0.75in 2.0in 0.75in},clip]{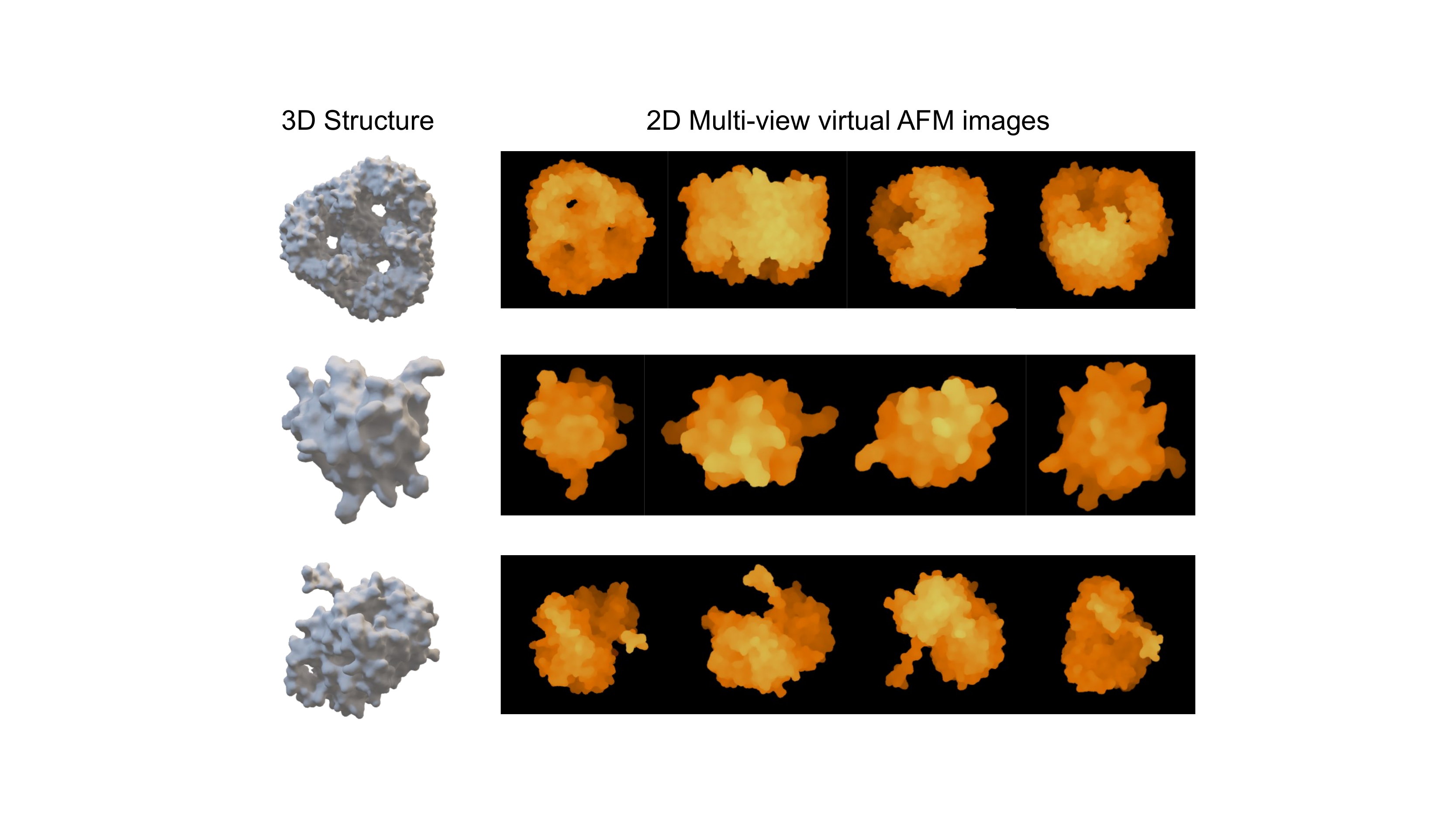}
    \caption{Showing the examples for virtual multi-view AFM images.}
    \label{fig:virtual_afm_ex}
\end{figure}


\end{document}